# 3D Human Pose Estimation for Free-form Activity Using WiFi Signals

YILI REN, Florida State University, USA
ZI WANG, Florida State University, USA
SHENG TAN, Trinity University, USA
YINGYING CHEN, Rutgers University, USA
JIE YANG, Florida State University, USA

WiFi human sensing has become increasingly attractive in enabling emerging human-computer interaction applications. The corresponding technique has gradually evolved from the classification of multiple activity types to more fine-grained tracking of 3D human poses. However, existing WiFi-based 3D human pose tracking is limited to a set of predefined activities. In this work, we present Winect, a 3D human pose tracking system for free-form activity using commodity WiFi devices. Our system tracks free-form activity by estimating a 3D skeleton pose that consists of a set of joints of the human body. In particular, we combine the signal separation and the joint movement modeling to achieve free-form activity tracking. Our system first identifies the moving limbs by leveraging the two-dimensional angle of arrival of the signals reflected off the human body and separates the entangled signals for each limb. Then, it tracks each limb and constructs a 3D skeleton of the body by modeling the inherent relationship between the movements of the limb and the corresponding joints. Our evaluation results show that Winect is environment-independent and achieves centimeter-level accuracy for free-form activity tracking under various challenging environments including the none-line-of-sight (NLoS) scenarios.

CCS Concepts: • **Human-centered computing** → **Ubiquitous and mobile computing systems and tools**.

Additional Key Words and Phrases: WiFi Sensing, Channel State Information (CSI), Commodity WiFi, Human Pose Estimation, Free-form Activity, 3D Human Skeleton

## 1 INTRODUCTION

In recent years, WiFi-based sensing is gaining increasing attention due to the prevalence of WiFi devices and their ability to sense the surrounding environments. Indeed, a variety of WiFi-based systems have been proposed to sense various human activities and objects, ranging from large-scale activities [39, 41, 64], indoor localization [28, 40, 61, 68, 76] to small-scale movements [3, 29, 52, 65], daily objects [25, 44, 55], and multi-person tracking [54, 66, 71]. For example, E-eyes [64] and WiFinger [52] are among the first work to distinguish various

Authors' addresses: Yili Ren, Florida State University, 1017 Academic Way, Tallahassee, FL, 32306, USA, ren@cs.fsu.edu; Zi Wang, Florida State University, 1017 Academic Way, Tallahassee, FL, 32306, USA, ziwang@cs.fsu.edu; Sheng Tan, Trinity University, One Trinity Place, San Antonio, TX, 78212, USA, stan@trinity.edu; Yingying Chen, Rutgers University, 671 Route 1, North Brunswick, NJ, 08902, USA, yingche@scarletmail.rutgers.edu; Jie Yang, Florida State University, 1017 Academic Way, Tallahassee, FL, 32306, USA, jie.yang@cs.fsu.edu.

types of daily activity and finger gesture based on the multi-class classification, whereas Liu *et al.* proposed the first generation vital signs tracking system with commodity WiFi [29]. More recent work has evolved towards constructing a 3D human pose that consists of a set of joints of the body at an unprecedented level of granularity [20]. However, existing WiFi-based 3D human pose tracking is limited to only a set of predefined activities as it relies on the pre-trained model of known activities. It thus cannot work well for free-form activities that were previously unseen by the system. In reality, there exists a variety of emerging Human-Computer Interaction (HCI) applications that demand the 3D human pose of free-form activity. For instance, Virtual Reality (VR) applications such as Fruit Ninja [14] require capturing the free-form movements of two arms of a player in 3D space. Meanwhile, art creation in VR needs to track the 3D free-form movements of two hands that simulate paintbrushes and color palettes [13]. Moreover, medical training in Extended Reality (XR) demands free-form motion tracking to enable trainees to learn about surgical operations by using hands and arms to interact with a 3D virtual human body [59, 60]. Additionally, the 3D free-form movement tracking can also enhance the control precision of existing smart home applications, such as continuous and precise thermostat temperature adjustment [38, 57].

Traditional systems in free-form human pose tracking mainly rely on computer vision technique that requires the installation of cameras in the environment or dedicated sensors that are worn/carried by the user. For example, 2D pose tracking can be achieved by using conventional RGB cameras [7, 17, 35], whereas 3D pose tracking can be done by leveraging depth or infrared cameras [31, 51, 73], such as Microsoft Kinect [32] or Leap Motion [58]. However, those systems cannot work in non-line-of-sight (NLoS) or poor lighting conditions and often involve user privacy concerns. The dedicated sensor-based approaches require a user to wear/carry various sensors such as gyroscope, accelerometer or inertial measurement unit (IMU) at each limb or joint [30, 50]. HTC VIVE [19] and Oculus Rift [34] are two examples of commercial products. These systems, however, can be inconvenient and cumbersome as they require the user's explicit involvement and incur the non-negligible cost.

In this work, we propose Winect, a skeleton-based human pose tracking system for free-form activity in 3D space using commodity WiFi devices. Winect does not rely on a set of predefined activities, thus can track free-form movements of multiple limbs simultaneously to enable novel HCI applications. It leverages the WiFi signals reflected off the human body for 3D pose tracking and thus works well under a non-line-of-sigh (NLoS) environment and does not require a user to wear or carry any sensors. Winect is environment-independent and provides centimeter-level tracking accuracy. It can also track the free-form activity of one or more users in the 3D space by utilizing multiple transceivers. In addition, the system could reuse existing WiFi infrastructure to facilitate large-scale deployment due to the prevalence of WiFi devices and networks.

The basic idea of Winect is to combine the signal separation and the joint movement modeling to enable 3D free-form activity tracking. In particular, we first develop a limb identification method that leverages the two-dimensional (2D) angle of arrival (AoA) of the signals reflected off the human body to infer the number of moving limbs and identify the limbs that are in motion. As the signal reflections from multiple limbs are linearly mixed at each antenna of the receiver [71], we separate the multi-limb signals based on the blind source separation (BSS) [1] and the input of the number of moving limbs. Once the signal reflections from each limb are separated, we can derive the position of each limb over time and infer the trajectory in 3D space with multiple transmitter-receiver pairs by leveraging phase changes of separated signals.

Next, we decompose the trajectory of each limb (e.g., the arm or leg) to the fine-grained trajectories of the joints (e.g., the wrist and elbow, or the ankle and knee) for 3D pose tracking. Specifically, we leverage the inherent relationship between the limb and the joints to construct a model that constrains the positions of corresponding joints for a given position of the limb. Existing kinematic models of limb joints [6, 11, 42] describe the position constraints of the joints for a given position of any point of the limb. The movement of the joints in a continuous and constrained space can be further utilized to optimize the positions of joints with respect to the positions of the limb over time.

We experimentally evaluate Winect system in a home environment (i.e., in both living room and bedroom) with various free-form activities conducted by different users. We evaluated our system under different numbers of moving limbs and users, various distances between the transmitter and receivers, non-line-of-sight environments, and different ambient interferences. The evaluation results show that our system can track 3D human pose with an average error of 4.6cm for various free-form activities. The main contributions of our work can be summarized as follows:

- We propose a 3D human pose tracking system that works for free-form activity by using commodity WiFi. The proposed system does not require a user to wear any sensors and works under NLoS scenarios.
- To enable free-form activity tracking, we separate the entangled signals from multiple limbs based on a blind source separation (BSS) formulation and decompose the movements of limbs to the joints by leveraging the kinematic model of limb joints.
- We conduct extensive experiments in various experimental settings. The evaluation results demonstrate that Winect achieves centimeter-level accuracy for free-form activity tracking under various challenging environments including the NLoS scenarios.

## 2 RELATED WORK

Existing work in human pose estimation can be divided into three categories based on the sensing techniques: computer vision-based, wearable sensor-based and RF sensing-based.

**Computer vision-based.** Both 2D and 3D human pose or skeleton estimation from images and videos are well-studied in the computer vision community. Due to the advancement of deep learning algorithms and the increasing availability of the annotated 2D human pose datasets, 2D pose estimation based on conventional RGB cameras has made tremendous progress [7, 35]. Meanwhile, significant research efforts have been devoted to 3D human pose estimation. For example, Sigal *et al.* [51] proposed an algorithm for 3D human pose tracking that uses a relatively standard Bayesian framework and VICON system. Pavllo *et al.* [36] introduced a fully convolutional architecture that performs temporal convolutions for accurate 3D pose prediction in videos. Mehta *et al.* presented VNect [31], a real-time method to capture the full 3D skeletal pose of a human in a consistent manner using a single RGB camera and the convolutional neural network (CNN). Kanazawa *et al.* [21] presented an end-to-end framework for recovering a full 3D mesh model of a human body from a single RGB image. Moreover, LiSense [26] achieves real-time 3D human skeleton reconstruction by analyzing shadows created by the human body from the blocked visible light. In addition, commercial off-the-shelf (COTS) products, such as Microsoft Kinect [32] and Leap Motion [58], leverage the depth or infrared cameras to track 3D human activities. However, computer vision-based approaches only work under the line-of-sight (LoS) scenario. Thus, their tracking performance would suffer if the target person is obstructed or under poor lighting conditions. Furthermore, these methods would incur non-negligible installation overhead and often involve user privacy concerns for daily users.

**Wearable sensor-based.** The wearable sensor-based approach requires a user to wear or carry various sensors. For example, by placing multiple sensors placed on each arm, existing work [9, 22] can track the movement of upper limbs. Some systems [45, 56] utilize multiple accelerometers that are attached to different parts of the user's body to reconstruct full-body motions. For example, SensX [8] leverages four inertial sensors to track acceleration and rotation of the human body for complex human motion tracking. Meanwhile, some research works track body movements with only a single sensor. For instance, RecoFit [33] automatically tracks repetitive exercises using a single inertial sensor on the arm. It can provide real-time and post-workout feedback by classifying multiple exercises and counting repetition. Shen *et al.* proposed MiLift [48], a workout tracking system that uses a smartwatch to accurately track both cardio and weightlifting exercises. ArmTrack [50] reconstructs movements of the entire arm using a single smartwatch. Similarly, a system like MUSE [49] can achieve slightly better localization accuracy compared to ArmTrack. Furthermore, ArmTroi [30] conducts arm skeleton tracking and activity recognition with higher accuracy and a reduced computational cost. However, the wearable sensor-based systems require the user to carry or wear one or more physical sensors, which could be inconvenient, intrusive, and cumbersome as it requires the user's explicit involvement.

**RF sensing-based.** To overcome the limitations of computer vision-based and wearable sensor-based systems, researchers have turned to RF sensing-based approaches for multi-limb movement tracking in recent years. Among them, some RF-based sensing systems [29, 41, 52, 64] leverage multiclass classification for coarse-grained activity or gesture recognition. For example, E-eyes [64] and WiFinger [52] is among the first work in WiFi human sensing to classify different daily activities and finger gestures respectively, whereas Liu *et al.* [29] proposed one of the first vital signs and figure movements tracking systems with commodity WiFi. On the other hand, recent RF-based sensing systems aim to achieve fine-grained pose tracking based on the trajectories of multiple joints. For example, RF-Capture [2] tracks the positions of the human limbs and body parts even when the person is fully occluded by the wall. Zhao *et al.* proposed RF-Pose [74] to achieve accurate 2D human pose estimation through the walls utilizing RF signals and a teacher-student network. Zhao *et al.* further proposed RF-Pose3D [75], a system that can infer 3D human skeletons from RF signals leveraging a novel convolutional neural network architecture. However, those systems all require specialized hardware (i.e., a carefully designed and synchronized T-shaped antenna array that contains a 4-antenna vertical array and a 16-antenna horizontal array) and the Frequency-Modulated Continuous Wave (FMCW) radio that repeatedly sweeps the band 5.46 to 7.24 GHz every 2.5 ms. Therefore, it is difficult for large-scale deployment due to the non-negligible cost and installation overhead. Person-in-WiFi [62] is a human pose estimation system that can reconstruct a 2D skeleton from WiFi signals. However, this system is limited to 2D tracking as it highly depends on the features (e.g., Part Affinity Fields [7] and Segmentation Masks [17]) that are only suitable for 2D scenes. Recently, WiPose [20] is proposed to construct a 3D human pose of many daily activities using WiFi devices. However, the number of poses that can be reconstructed is limited to only a set of predefined activities in the training phase. It thus cannot work well for free-form activities that unseen by the system. In our work, propose a 3D human pose tracking system that works for free-form activity, which could enable novel HCI applications that require tracking free-form movements of multiple limbs.

## 3 SYSTEM DESIGN

In this section, we discuss the design challenges, the system overview, the system flow and its core components.

### 3.1 Design Challenges

Intuitively, we can quantify the movement of the human body by analyzing the Channel State Information (CSI) extracted from the WiFi signals that are perturbed by the moving limbs of the user [63, 65, 69]. The extracted complex-valued CSI measurement could represent both the phase and the amplitude of the received wireless signal. The distance of the movements thus can be quantified by calculating the phase change of the signals reflected off the human body. Further, the direction of the movement can be quantified by the rotation direction of CSI in the complex plane. Specifically, when a limb moves away from a transmitter-receiver pair, the length of the reflected path increases, which leads to the clockwise rotation of CSI in the complex plane and vice versa. While the intuition is simple, there are significant challenges to accurately tracking 3D free-form activities that involve multiple limb movements.

**Dealing with multiple moving limbs.** Existing works have achieved fine-grained tracking for single hand [69] or finger [53, 65] movements using commodity WiFi. However, when multiple limbs are moving simultaneously, the perturbed WiFi signals will be mixed together. It is difficult to track each limb without proper signal separation. Although existing work [54] proposes to splice multiple channels together to increase the bandwidth and improve the distance resolution for signal separation of multiple user tracking, the granularity is still not enough to separate movements of nearby and closely spaced multiple limbs of the same person.

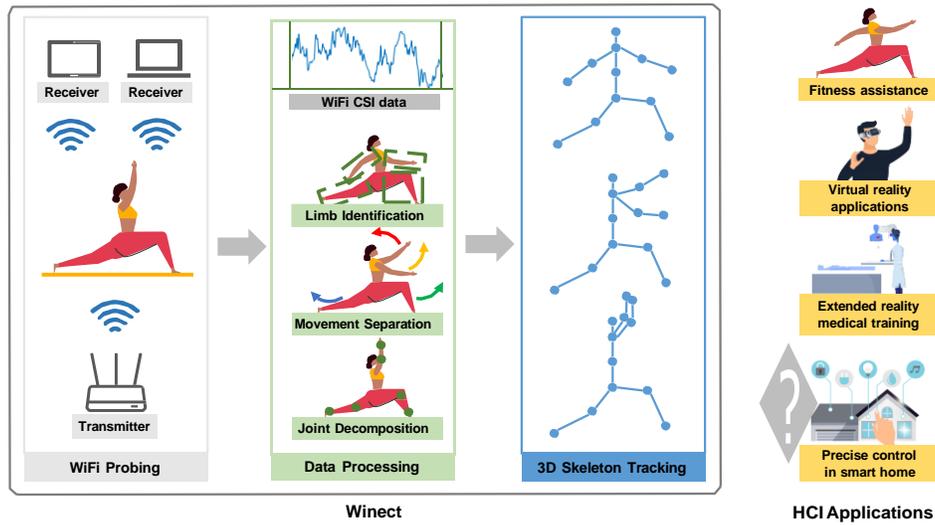

Fig. 1. System overview.

**Moving Limb identification.** To facilitate the 3D multi-limb and free-form tracking, our system needs to know the information of how many limbs are moving simultaneously and the specific limbs that are in motion. However, it is less desirable to manually input that information to the system during the initialization or tracking phase. To improve the useability of the system, it is essential for the system to identify the moving limb information automatically, which could be challenging in 3D pose tracking.

**Free-form activity tracking.** The state-of-the-art WiFi-based 3D human pose tracking is designed based on the training of a set of predefined activities [20]. Particularly, it leverages the deep neural network to reconstruct 3D poses for a set of typical daily activities. It is, however, hard to track free-form activities that were previously unseen during the training phase. How to enable free-form activity tracking without predefined activity training is still a challenge for WiFi-based 3D pose estimation.

### 3.2 System Overview

The key idea of our proposed 3D free-form activity tracking is to identify and separate the reflected signals from multiple moving limbs and decompose the movements of each limb to the corresponding joints. In particular, we consider a real-life scenario as shown in Figure 1, in which the user's activities are sensed by the WiFi signals emitted from multiple commodity WiFi devices. When the user is performing multi-limb free-from activities, the signal reflected off the human body will constructively or deconstructively combine with both the direct propagated signals and the signal reflected from the environments. Multiple receivers will then capture the combined signals at each antenna.

Our system once accepted the received signal as input, it extracts CSI measurements and identifies the number of moving limbs and the specific limbs that are in motion by leveraging the 2D AoA of the reflected signals. We next separate the reflected signals for different moving limbs based on a blind source separation (BSS) formulation. With the separated signals for each limb, we can further infer the positions or the trace of the individual limb by calculating the signal path length change based on the phase changes. Lastly, our system decomposes the inferred limb traces into fine-grained trajectories of the joints based on the kinematic model of limb joints, which depicts the inherent relationship between the position of the limb and its corresponding joints.

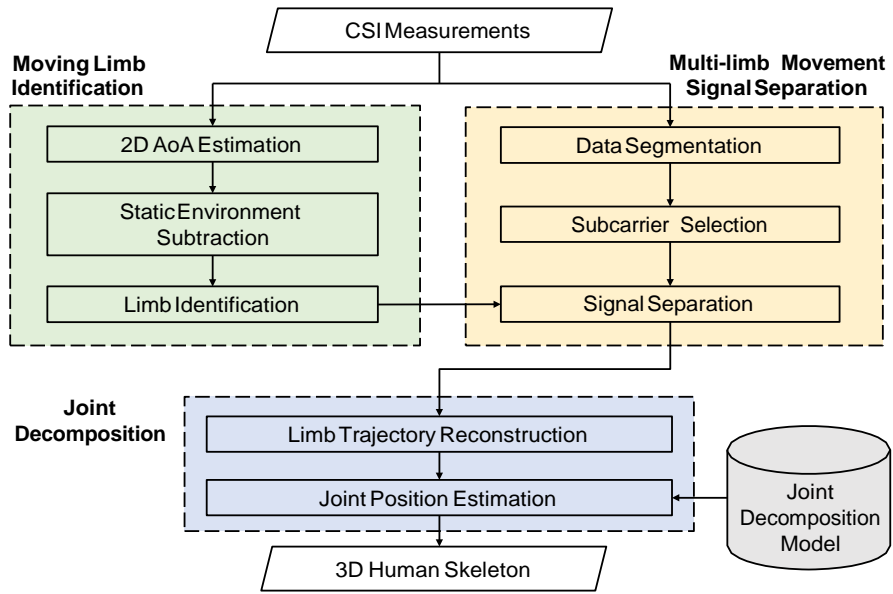

Fig. 2. System flow.

By utilizing both the multi-limb movement signal separation and the joint decomposition, our system can achieve multi-limb free-form tracking in 3D space. Such a system can support a wide range of emerging HCI applications that require the tracking of 3D free-form body movements, for example, in the fitness assistance system. Moreover, VR applications can utilize the free-form movement information for gaming and art creation. Such information can also be utilized to facilitate the extended reality medical training as well as in smart home applications, as shown in Figure 1.

### 3.3 System Flow

The system flow is illustrated in Figure 2, which consists of three major components: *Moving Limb Identification*, *Multi-limb Movement Signal Separation* and *Joint Decomposition*. Our system first performs CSI measurements collection, in which a single WiFi transmitter continuously sends out probe packets and multiple WiFi receivers extract CSI measurements from the received packets.

Winect first employs *Moving Limb Identification* to pre-process the captured CSI measurements and estimate the two-dimensional angle of arrival (i.e., the elevation angle and azimuth angle) of the multipath signals under both the static scenario without limb motion and the dynamic scenario where the limbs are moving. We then analyze the signal power change according to the derived 2D AoA (i.e., azimuth-elevation) spectrum in 3D space. Next, our system identifies the number of moving limbs and the specific limbs that are in motion. This information is then fed into the multi-limb movement signal separation component.

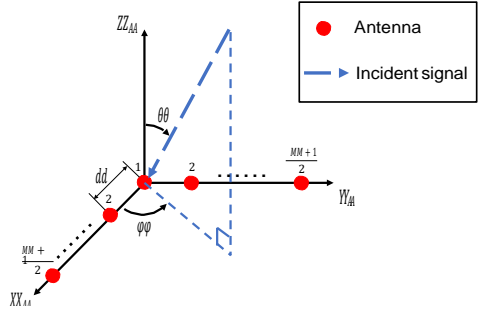

Fig. 3. The L-shaped array configuration used for the joint azimuth and elevation estimation.

Next, our system conducts *Multi-limb Movement Signal Separation* by first segmenting CSI measurements and selecting the subcarriers with high sensitivity to limb movements. Then, with the number of moving limbs identified from the previous step, we are able to separate the multi-limb movement signal by solving the formulated blind source separation (BSS) problem. After that, our system will calculate the limb position based on the path length change of the separated signals. We then stitch the limb positions from two adjacent segments that belong to the same limb to reconstruct the trajectory of the limb.

After we obtain the positions and trajectories of each limb, they go through *Joint Decomposition* component to further decompose to the joint positions by leveraging a kinematic model of limb joints. It is done by modeling the relationship between the position of the limb and the corresponding joint in 3D space. Then, the 3D human skeleton with multiple joints can be inferred. It is worth noticing that by combining both *Multi-limb Movement Signal Separation* and *Joint Decomposition*, our system achieves 3D multi-limb tracking for free-form activity.

### 3.4 Moving Limb Identification

In this section, we customize the 2D AoA estimation method leveraging L-shaped antenna array, spatial diversity in transmitting antenna and frequency diversity of OFDM subcarriers to enhance the resolution. We also develop a method to subtract the impact of the static environment for multi-limb identification.

*3.4.1 Preliminary of 2D AoA Estimation.* Traditional AoA estimation focuses on inferring the angle of arrival of the received signal in one dimensional with linear antenna array [24, 27]. It is however insufficient to identify the direction of the incoming signal in 3D space. In this work, we extend the traditional 1D AoA estimation to 2D AoA estimation so as to facilitate activity tracking in 3D space. In particular, we derive both the azimuth and elevation angles of the signal by leveraging an L-shaped antenna array at the receiver [16]. As shown in Figure 3, we assume the L-shaped antenna array separated (i.e., a half wavelength) at the receiver. It is worth noticing that the L-shaped array antenna coordinate system is different from the tracking coordinate system in this work. Then, we can formulate the 2D AoA estimation based on the MUSIC algorithm [47] and the L-shaped antenna array at the receiver.

*3.4.2 Improved 2D AoA Estimation.* As commodity WiFi devices are only equipped with a limited number of antennas (e.g., Intel 5300 card has up to only 3 antennas), this greatly limits the resolution of AoA estimation. To further improve the 2D AoA estimation, we propose to leverage both spatial diversity in transmitting antennas and frequency diversity of OFDM subcarriers of the WiFi channel on both two subarrays of the L-shaped antenna array. Both these two diversities can introduce phase shifts and can improve the resolution of AoA estimation. Therefore, we utilize CSI measurements of the WiFi signals across all OFDM subcarriers and transmitted from multiple transmitting antennas to generate a large number of sensing elements. In our implementation, for each subarray, we have 2 receiving antennas, 3 transmitting antennas, and 30 subcarriers. Thus, there are 2 × 3 × 30 sensing elements for each axis. They contain enough information that allows our system to jointly estimate 2D AoA, AoD (angle of departure), and ToF (time of flight) simultaneously. Those results can be combined to improve the resolution of 2D AoA estimation.

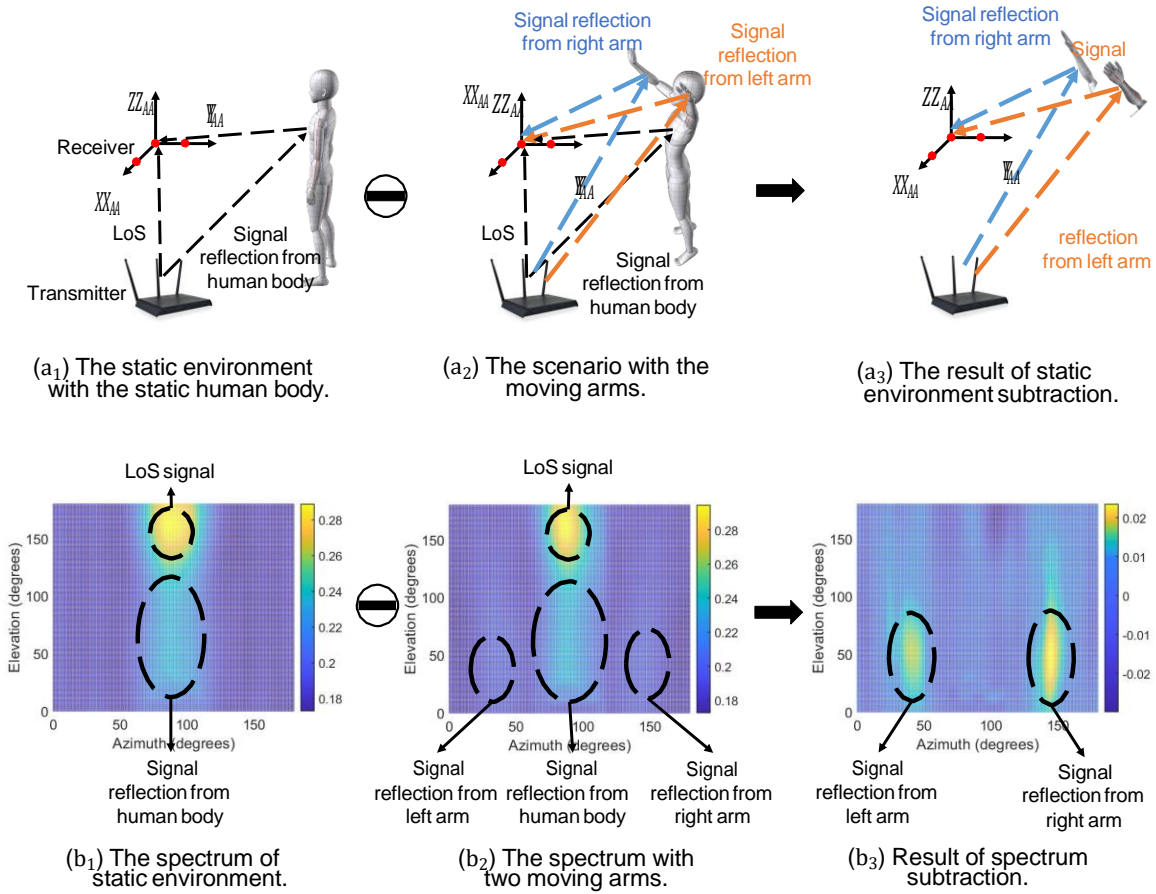

Fig. 4. Illustration of static environment subtraction in azimuth-elevation spectrum.

*3.4.3 Static Environment Subtraction and Limb Identification.* Inspired by the spectral subtraction [4, 5] which is done by the subtraction of the average noise spectrum estimation from the noisy signal spectrum, we identify each moving limb by conducting static environment subtraction [54] similarly and analyzing the signal power change in the azimuth-elevation power spectrum. In order to remove the impact of the static environment, we first derive the 2D AoA spectrum of the environment. By subtracting the static environment, we can extract the signal reflected from the moving limb, which is independent of the environment. For example, as shown in Figure 4($a_1$), a user is standing along the axis and perpendicularly faces to the coordinates in a static

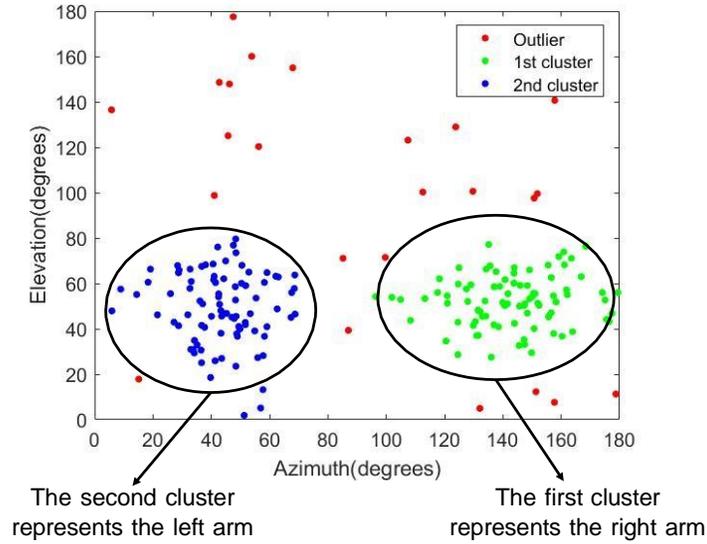

The second cluster represents the left arm

The first cluster represents the right arm environment. The propagation paths are all static paths including the LoS path, signal reflections from the static environment (including the static human body). The corresponding azimuth-elevation power spectrum is shown in Figure 4($b_1$) where the x-axis represents the azimuth angle, the y-axis represents the elevation angle and the color represents the power. Since there are no movements from the user, the power spectrum only contains the signal reflections from the static environments and the LoS propagation.

Fig. 5. Limb identification result.

Then, when the user is moving two limbs (e.g., the left arm and right arm) as demonstrated in Figure 4($a_2$), two additional dynamic paths are introduced by the motions of the left and right arm. Figure 4($b_2$) is the corresponding power spectrum that contains signal reflections from both arms, the static environments and the LoS propagation. Then, we perform static environment subtraction where the normalized azimuth-elevation spectrum under the dynamic scenario is subtracted by the normalized spectrum of the static environment. The result is shown in Figure 4($b_3$). After the subtraction, we can obtain the azimuth-elevation spectrum that containing the signal propagation mainly affected by the moving limbs. As we can easily observe in Figure 4(b3) and Figure 4($a_3$), there are two major signal reflection components that represent the left arm and the right arm, respectively.

Next, we detect the peak values in the resulting azimuth-elevation spectrum to identify all the limbs that are in motion. Intuitively, the number of peaks is corresponding to the number of moving limbs while we can further infer the specific limb based on the peak position in the azimuth-elevation spectrum. Here, we conduct the identification process using multiple random CSI packets (e.g., 100 packets) to mitigate the random error introduced by a single packet. In particular, Winect uses a non-parametric clustering method: Density-based spatial clustering of applications with noise (DBSCAN) algorithm [10] to cluster the peaks without prior knowledge of the peak number. The limb identification result is shown in Figure 5. We can observe that there are two clusters in total therefore two moving limbs are detected. Then, we can easily calculate the average position of each cluster and pinpoint the limb according to the average peak position in the azimuth-elevation spectrum. Under the system setup illustrated in Figure 4, if the azimuth of the peak position is less than 90 degrees and the elevation of the peak position is also less than 90 degrees, such cluster will be identified as the left arm. If the azimuth of the peak position is less than 90 degrees and the elevation of the peak position is greater than 90 degrees, such a cluster would be identified as the left leg. Similarly, the other two limbs can also be identified respectively. Note that the obtained number of moving limbs will be taken as prior knowledge of the later signal separation section.

### 3.5 Multi-limb Movement Signal Separation

*3.5.1 Data Segmentation and Subcarrier Selection.* In order to facilitate the multi-limb movement signal separation and limb trajectory reconstruction, we first need to perform data segmentation. In particular, we divide CSI measurements into a series of 0.1-second segments. After segmentation, each CSI segment can be further separated into static components and dynamic components which correspond to static paths and dynamic paths of the multipath. Dynamic components consist of the signals reflected from each moving limb and static components consist of the LoS signal and signals reflected from the static surrounding environments.

Each subcarrier experiences different multipath fading effects and the CSI measurements of different subcarriers have various sensitivity to the limb movement. Thus, it is important to select the most sensitive subcarrier for each antenna pair to further improve tracking accuracy. In particular, we first calculate the movement energy ratio [71]. It is done by taking fast Fourier transform (FFT) of the CSI amplitude and dividing the energy sum of all FFT bins in the frequency range of limb movement (e.g., 0-20 Hz) by the energy sum of all FFT bins. Then, for each subcarrier, we average the movement energy ratio of all antenna pairs. Lastly, we select the subcarrier that has the maximum movement energy ratio for each segmented CSI measurement. This is because a larger movement energy ratio means the selected subcarrier is more sensitive to the limb movement.

*3.5.2 Signal Separation.* In this section, we show that the multi-limb movement signal separation can be modeled as a Blind Source Separation (BSS) problem. Then, we can solve the BSS problem by leveraging the Independent Component Analysis (ICA) with the identified number of moving limbs in the previous step.

**Independence and Non-Gaussianity of Sources.** The BSS problem can be solved using ICA given that all the sources are independent, non-Gaussian, and combined linearly [46]. To verify such an assumption, we first examine the independence of the possible sources (i.e., movements of different limbs). Here, we ask different users to perform free-form activities with multiple limbs and capture the movement of each limb with Kinect 2.1 to serve as the ground truth. Next, we divide the captured movements into non-overlap periods. Then, we compute the correlation between two real limb movements during each period and repeat the process for all users. Figure 6 shows the correlations between different limb movements and it is easy to observe that all correlations values are smaller than 0.08. Therefore, we can assume that the movement of different limbs is independent of each other. To demonstrate the non-Gaussianity of the sources, we show the distributions of the limb movements in Figure 7. Here, the x-axis represents the average position of all joints of the limb during movements and the y-axis represents the amount. We can observe that the distributions are non-Gaussian.

**Linearity of Observations.** Next, we will demonstrate that the captured CSI measurements are linear mixtures of the limb movement signal.

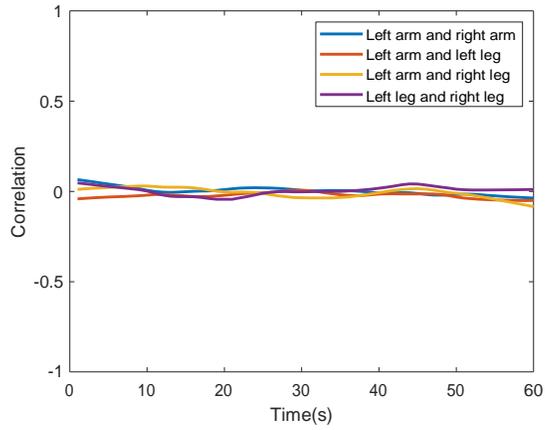

Fig. 6. Correlation between two different limb movements.

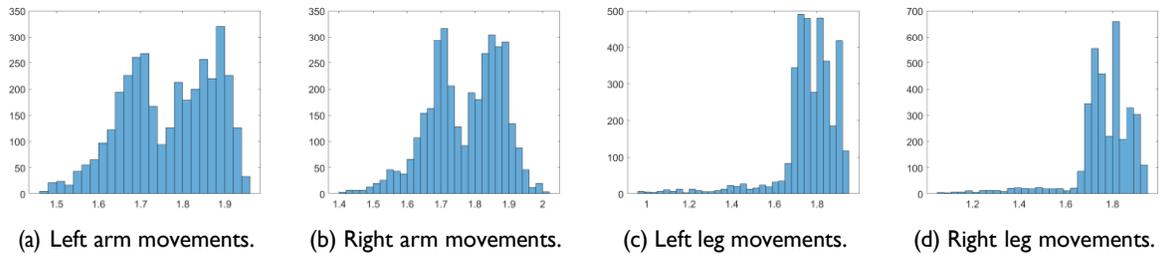

(a) Left arm movements.  (b) Right arm movements.  (c) Left leg movements.  (d) Right leg movements.

Fig. 7. Illustration of non-Gaussian distributions of the limb movements.

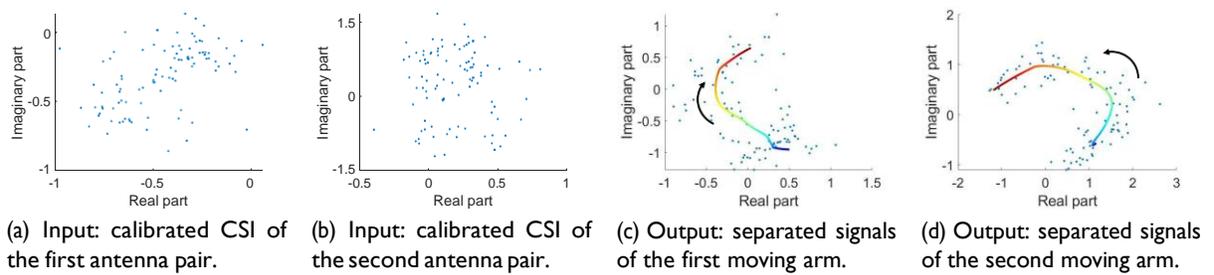

(a) Input: calibrated CSI of the first antenna pair.  (b) Input: calibrated CSI of the second antenna pair.  (c) Output: separated signals of the first moving arm.  (d) Output: separated signals of the second moving arm.

Fig. 8. An example of separating movement signals of two moving arms.

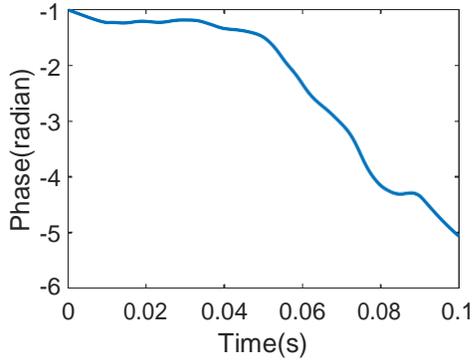
(a) Phase change of the separated signal of the first arm.

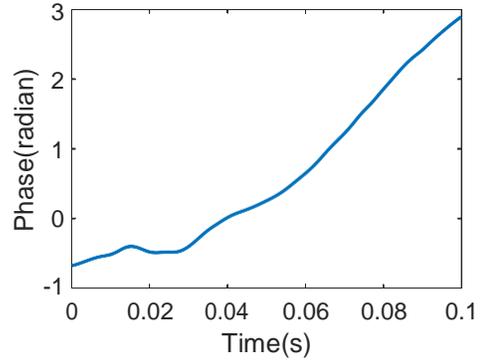
(b) Phase change of the separated signal of the second arm.

Fig. 9. Phase changes of separated signals.

**Separation Implementation and Path Length Change Calculation.** The output of Section 3.4 provides the number of moving limbs which will be utilized as the number of the sources for the ICA algorithm. Then, we further select the calibrated CSI segments from antenna pairs to feed into ICA as input. Note that we only select the top antenna pairs that have a higher movement energy ratio discussed in Section 3.5.1. Here, we utilize the RobustICA algorithm [70] which can work for complex-valued signals and has a high convergence speed.

Figure 8 shows an example of multi-limbs movement signal separation for a single CSI segment (e.g., 0.1s) using RobustICA when the user is moving both arms. Here, the dots represent the time series data and the colored lines represent the denoised data. Figure 8(a) and (b) depict the calibrated CSI of two different antenna pairs in the complex plane. Combining with the number of sources, they will be fed into RobustICA as input. The results are shown in Figure 8(c) and (d), in which the points are the time series samples in the complex plane and the colored lines are the smoothed version by using a Savitzky-Golay filter. In Figure 8(c), it is easy to observe that the separated signals of the first arm rotate clockwise in the complex plane which indicates the first limb moves away from the transmitter-receiver pair. Meanwhile, we can observe that the second arm moves towards the transmitter-receiver pair as shown in Figure 8(d).

After that, our system will calculate the path length change based on the signal rotation in the complex plane. Thus, as shown in Figure 9(a), the phase change caused by the first moving arm is 3.25 radians, which can be translated to a path length change of 2.9cm. Similarly, Figure 9(b) shows the second moving arm induces a phase change of 3.51 radians which can be translated to a path length change of 3.2cm.

### 3.6 Joint Decomposition

*3.6.1 Limb Position Estimation and Trajectory Reconstruction.* After multi-limb movement signal separation, we are able to infer the movement direction and path length change of the limb from a single transmitter-receiver pair. However, it can only estimate the limb position and reconstruct the trajectory of the limb in 1D. In order to achieve multi-limb tracking in 3D space, we utilize three transmitter-receiver pairs and build the tracking coordinate system shown in Figure 10. Here, the transmitter is at the origin of the coordinates, respectively. Therefore, we are able to repeat the previous steps for each transmitter-receiver pair to enable multi-limb tracking in 3D space. Given that the distance between each transmitter and receiver is known, we can localize all potential limb positions by intersecting three ellipsoids with foci at the corresponding transmitter-receiver pair. It is worth noticing that we set the midpoint of the forearm or the midpoint of the lower leg of the initial skeleton (i.e., standing with arms naturally hanging down) as the initial positions for trajectory reconstruction.

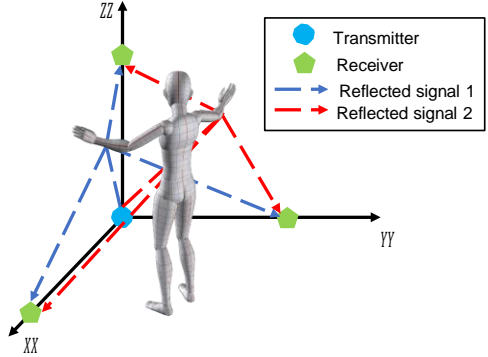 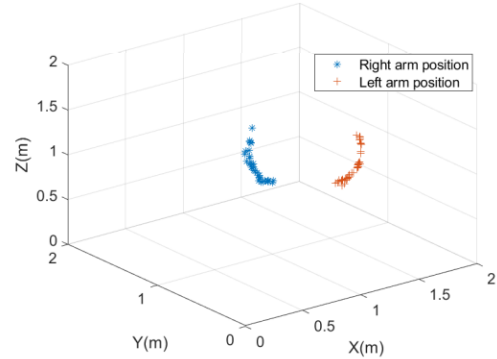

Fig. 10. 3D tracking with three transmitter-receiver pairs.   Fig. 11. Reconstructed trajectories of two moving arms.

Specifically, we assume there are $\alpha$ path length changes for each of the $\beta$ receivers after multi-limb movement signal separation. Here, we stitch the positions from two adjacent segments that belong to the same limb from each receiver to reconstruct the trajectory of the limb. Since we could use any of the $\alpha$ obtained path length change from all $\beta$ receivers to calculate the limb position, there will be $\alpha^\beta$ potential limb positions in total. The insight of trajectory reconstruction is that the position of each limb changes slowly within a short amount of time (i.e., a single CSI segment) and it is significantly different from other potential limb positions. Therefore, it is possible to calculate all $\alpha^\beta$ potential limb positions, and the new position for each limb will be the one that satisfies the temporal and spatial constraints. By repeating this process, we are able to reconstruct the trajectory of all the limbs. Figure 11 shows the reconstructed trajectories of two moving arms corresponding to the movements shown in Figure 10. Note that both $\alpha$ and $\beta$ are small values and will not lead to a large increase in computational complexity.

*3.6.2 Joint Position Estimation.* Even after we obtain the free-form trajectory of each limb, the granularity is still not enough to support many emerging HCI applications (e.g., fitness assistance). Thus, it is important to infer the trajectories of the corresponding joints to provide more detailed positions in 3D space. Existing work [50] can track the positions of the joint by formulating a mathematical model. However, it requires the user to attach the sensors on the limb (e.g., the wrist) which can not simply be applied to our system. In our work, by carefully analyzing the existing kinematic model [6, 11, 42] of the arm and the leg, we find that the arm and leg joints have certain limitations for their range of motion (as shown in Figure 12) and they will also constrain the activities. This inspires us to utilize deep learning to model the inherent relationship between the position of the limb and the corresponding joints. In this work, we denote a set of discrete position points (i.e., the position of the limb or the joint) in the range of motion as a point cloud which is shown in Figure 13 and we can easily build the point clouds for both limb positions and joint positions. If the point cloud is dense enough, almost all daily life free-form activities can be represented as the paths formed by points in the point cloud as shown in Figure 13.

Thus, we could construct a model using the deep learning algorithm that can learn the relationship between the point cloud of limb positions and the point cloud of joint positions. By using the point cloud, the model training is not based on the signal pattern and thus is environment-independent. Therefore, building such a model is easy and simple for the users.

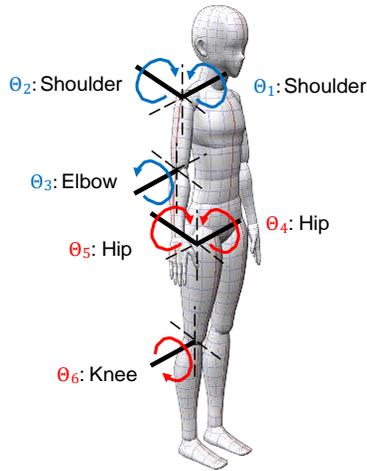

Table 1. Range of motions for each joint angle.

| Joint angle | Min. Value | Max. Value |
|---|---|---|
| $\Theta_1$ | −60° | 180° |
| $\Theta_2$ | 0° | 120° |
| $\Theta_3$ | 0° | 145° |
| $\Theta_4$ | −10° | 50° |
| $\Theta_5$ | 0° | 45° |
| $\Theta_6$ | 0° | 130° |

Fig. 12. Human joints with possible angular rotations.

**Point Cloud Construction.** Inspired by the robotics studies, we consider 3 degrees of freedoms (DoFs) [37] of the joints of a single arm and 3 DoFs of the joints of a single leg. Figure 12 shows the right side of DoFs of human joints, while the left side has the same joints. In particular, $\Theta_1$ and $\Theta_2$ are the 2 DoFs on the shoulder joint, which correspond to flexion/extension and abduction/adduction rotation of the shoulder, respectively. $\Theta_3$ is the only DoF on the elbow joint, and it represents flexion/extension on the elbow. The DoFs of the leg can be similar to DoFs of the arm. In order to derive the point cloud for each limb and each joint, we referred to some kinematic papers [6, 11, 42] and summarized the average range of motion for each joint angle in Table 1. It is worth noting that, to avoid overlapping of limbs and make it easy to separate signals, we set the minimum value of $\Theta_2$ and $\Theta_5$ to 0 degrees in our system.

We then discuss in detail how to construct the point cloud. A user can construct the limb point cloud as shown in Figure 13(a) according to Algorithm 1. We do not specify $\Theta_3$ and $\Theta_6$ in the algorithm and the user can freely adjust them to comfortable degrees during the activity. $\Delta\Theta$ is the step size of the change of $\Theta_2$ and $\Theta_5$. A smaller step size will result in a denser point cloud and a more accurate trajectory. However, it will lead to a longer time to construct the point cloud. Thus, we set $\Delta\Theta$ to 3° empirically. For joint point cloud, we use Kinect 2.0 to record the ground truth of positions of joints. Figure 13(b) shows an example of the point cloud for joints of the limb. Our goal is to build a model which predicts the joint positions for a given limb position.

**Construction of Joint Decomposition Model.** To build such a model, Winect leverages a ResNet [18], which is a widely-used network architecture. In addition, its convolutional neural network can extract spatial features well. Specifically, the ResNet will take limb point cloud and joint point cloud as input and outputs the predicted positions of multiple joints. We consider the problem of joint localization as a regression problem that

**Algorithm 1:** Point Cloud for Limb Positions
  **Input:** CSI measurements
  **Output:** LimbPointCloud
1 WiFi probing;
2 Perform Section 3.4 Moving Limb Identification;
3 Perform Section 3.5 Multi-limb Movement Signal Separation;
4 $\Theta_2 = 0°$; $\Theta_5 = 0°$;
5 **if** *left/right arm* **then**
6    **if** $\Theta_2 \leq 120°$ **then**
7       Move arm from $\Theta_1 = -60°$ to $\Theta_1 = 180°$;
8       Perform Section 3.6.1 Limb Position Estimation and Trajectory Reconstruction;
9       LimbPointCloud.Add(left/right arm position);
10      $\Theta_2 = \Theta_2 + \Delta\Theta$;
11    **end**
12 **end**
13 **if** *left/right leg* **then**
14    **if** $\Theta_5 \leq 45°$ **then**
15       Move leg from $\Theta_4 = -10°$ to $\Theta_4 = 50°$;
16       Perform Section 3.6.1 Limb Position Estimation and Trajectory Reconstruction;
17       LimbPointCloud.Add(left/right leg position);
18      $\Theta_5 = \Theta_5 + \Delta\Theta$;
19    **end**
20 **end**

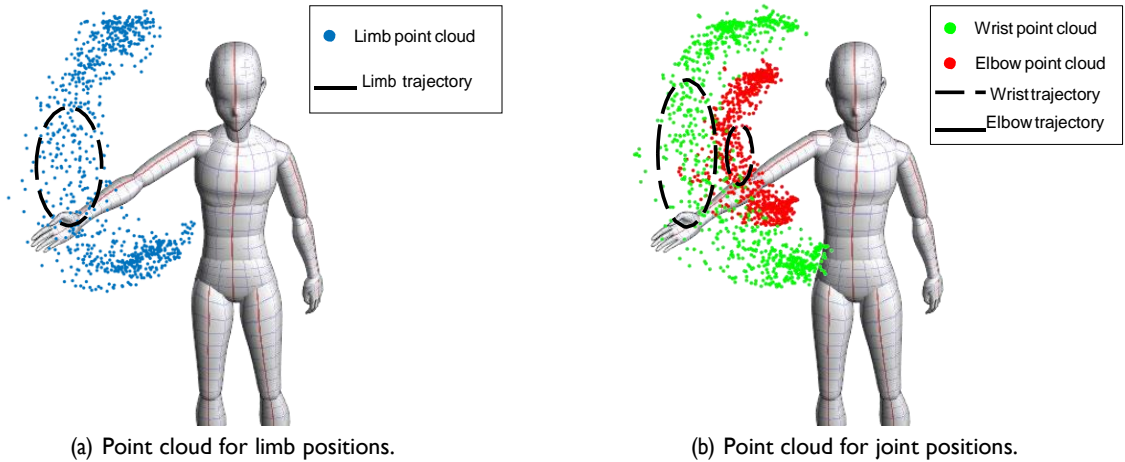

(a) Point cloud for limb positions.  (b) Point cloud for joint positions.

Fig. 13. Examples of point clouds for both the limb and joints.

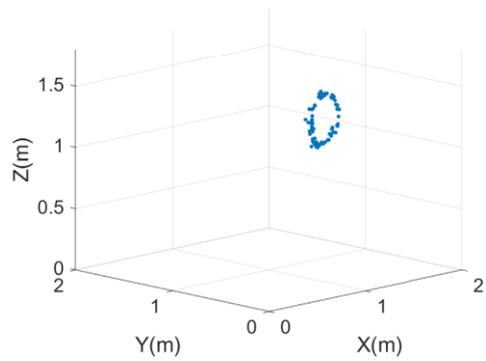
(a) Trajectory of the limb before decomposition.

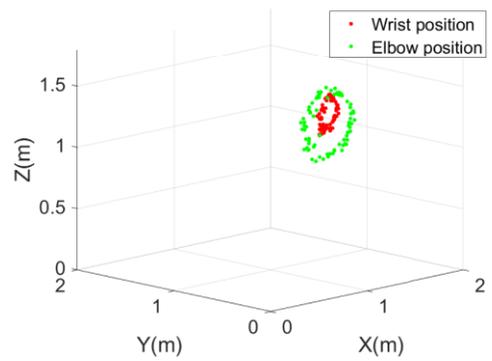
(b) Trajectories of the joints after decomposition.

Fig. 14. An example of joint decomposition.

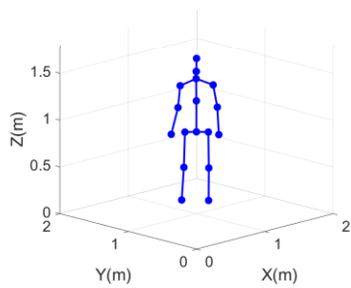
(a) Lifting both arms (start).

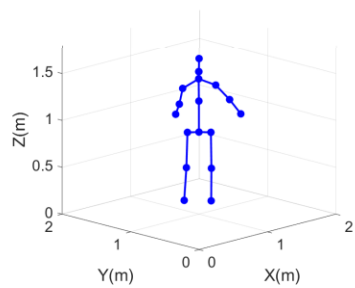
(b) Lifting both arms (middle).

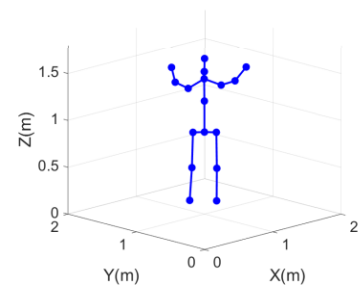
(c) Lifting both arms (end).

Fig. 15. Continuous 3D human skeleton for lifting both arms.

can estimate the relationships between limb positions and joint positions. Training the neural network aims to minimize the average Euclidean distance error between predicted joints and the ground truth.

:

**Joint Decomposition and Skeleton Reconstruction.** Once the model is trained for multiple limbs, it can predict the positions of joints for each given position of each limb. Thus, once we have the time series limb positions of free-form activities, we can construct the trajectories of multiple joints for the corresponding activities. Figure 14(a) illustrates a person is performing free-form activities (e.g., drawing circles) with the arm. Then, the time series limb positions can form the limb trajectory and can be taken as input to the trained joint decomposition model. Next, the model outputs the decomposition results for joints positions as shown in Figure 14(b). They are two circles consist of positions of the wrist and the elbow respectively. Moreover, we can reconstruct the continuous 3D human skeleton with the time series joint positions. For example, Figure 15 shows three statuses (i.e., the start, middle and end) of the 3D skeleton of the free-form activities with both arms. Note that if the skeletal structure of the person in terms of bone length is available, we can construct the initial skeleton. If this prior knowledge is not available, we can estimate the skeletal structure based on the height of the person or utilize a default skeletal structure.

## 4 PERFORMANCE EVALUATION

In this section, we evaluate the performance of our system using the commodity WiFi devices in different environments under various scenarios.

### 4.1 Experimental Setup

**Devices.** We conduct all experiments with five laptops (i.e., one transmitter, four receivers). Each laptop runs Ubuntu 14.04 LTS and is equipped with Intel 5300 wireless NIC connected with three omnidirectional antennas. Linux 802.11 CSI tools [15] are used in our experiments to extract CSI data for both limb identification and multi-limb tracking. We set the frequency of the WiFi channel to 5.32 GHz with a bandwidth of 40 MHz. If not specified, the packet rate is set at 1000 packets per second. In order to obtain the ground truth of the movement of multiple limbs, we utilize a Microsoft Kinect 2.0 [32] and the sampling rate of Kinect 2.0 is set at 10 Hz. We synchronize the local clock on all the receivers and the Kinect 2.0.

**Environments.** We conduct experiments in both the living room and bedroom environments. Figure 16(a) shows the top view of the deployment of devices in each environment. The subject is facing the transmitter while performing free-form activities. We place a Microsoft Kinect 2.0 to record the whole body movement of the subject. Figure 16(b) illustrates the 3D perspective of the deployment of the devices. Specifically, Receiver 1, Receiver 2 and Receiver 3 are placed on the three axes of the tracking coordinate system respectively for 3D tracking. The transmitter is at the origin of the coordinates and the distance between the transmitter and each receiver (i.e., R1, R2 and R3) is 2 meters. Note that Receiver 4 is placed at a distance of

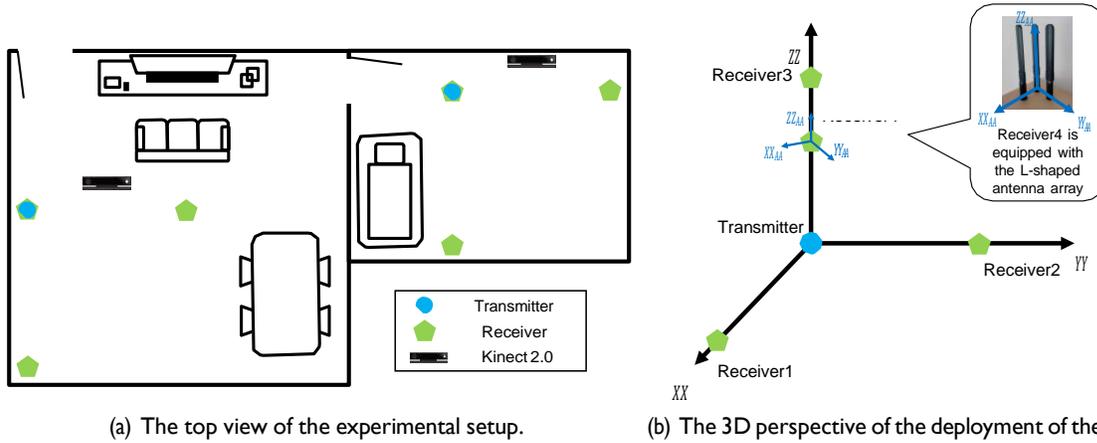

(a) The top view of the experimental setup.  (b) The 3D perspective of the deployment of the devices.

Fig. 16. Illustration of the experimental setup.

1.1 meters to the transmitter and equipped with the L-shaped antenna array which forms the L-shaped array antenna coordinate system discussed in Section 3.4 for moving limb identification.

**Model Setting.** In order to construct the joint decomposition model, we ask volunteers to conduct free-form daily life activities in terms of Table 1 and Algorithm 1. In order to cover the range of motions for different joints as much as possible and build a dense point cloud, each volunteer performs various activities for at least 20 minutes and we collect more than 24,000 points using CSI measurements to build the point cloud for the limbs. Kinect 2.0 synchronously records joint positions to build the point cloud for joints. The point clouds are dense enough to reconstruct trajectories for most free-form daily life activities. After that, we build the joint decomposition model using all the point clouds data.

In particular, we leverage the ResNet with 18 layers including 17 convolutional layers and one fully connected layer. After each convolutional layer, we add a batch normalization layer. We also utilize Rectified Linear Unit (ReLU) activation functions after each batch normalization layer to add non-linearity to the model. To prevent the model from overfitting, we set the dropout rate as 0.1.

**Data Collection.** In total, there are 6 volunteers (3 males and 3 females) of various heights who participate in the experiment. To evaluate the system performance, each participant is asked to conduct one-limb and multi-limb free-form daily life movements in the range of motions specified in Table 1 at least for 20 minutes. We segment the movements into a series of 8-second pieces and thus each 8-second piece is considered as a free-form activity. Therefore, there are over 900 activities in total. Note that the data are collected on different days and in different environments.

**WiPose.** WiPose [20] is the state-of-the-art deep learning model for 3D human pose construction for a set of predefined activities using commodity WiFi. WiPose combines four-layer Convolutional Neural Networks (CNNs) and three-layer Long Short Term Memory networks (LSTMs). Then it takes the initial skeletal structure and the learned features from LSTM as input for the forward kinematics layer. In this paper, we reproduce the deep learning model proposed in WiPose for comparison. We compare WiPose with our system with the predefined activities that were evaluated in WiPose paper including lifting left/right hand for 90/180 degrees, lifting left/right leg for 45 degrees, lifting both hands for 90/180 degrees, lifting left/right hand and left/right leg for 45 degrees and lifting both hands and left/right leg for 45 degrees. Each volunteer is asked to conduct each activity for two minutes. We also compare WiPose with our work for free-form activities in our evaluation.

Table 2. Joint localization errors (unit:*cm*).

| Joints | LElbow | LWrist | RElbow | RWrist | LKnee | LAnkle | RKnee | RAnkle | Overall |
|---|---|---|---|---|---|---|---|---|---|
| WiPose-predefined | 4.9 | 6.1 | 5.2 | 5.8 | 3.5 | 4.0 | 2.1 | 2.5 | 4.3 |
| WiPose-free-form | 13.7 | 24.6 | 13.2 | 20.6 | 12.0 | 14.5 | 13.4 | 15.9 | 16.0 |
| WiPose | 9.3 | 15.4 | 9.2 | 13.2 | 7.8 | 9.3 | 7.8 | 9.2 | 10.1 |
| Winect | 4.2 | 5.1 | 4.8 | 4.9 | 4.1 | 4.4 | 4.3 | 4.7 | 4.6 |

**Metrics.** To evaluate the performance of our system on multi-limb tracking, we utilize the joint localization error, which is defined as the Euclidean distance between the predicted joint location and the ground truth.

### 4.2 Overall Performance

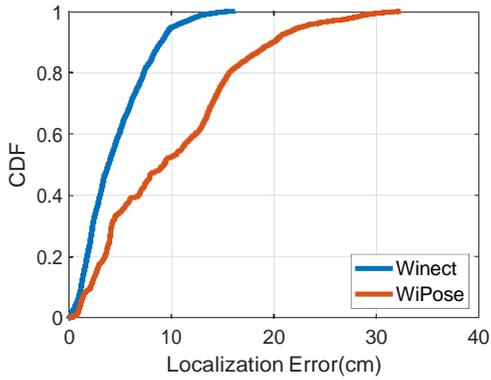
Fig. 17. Overall tracking error.

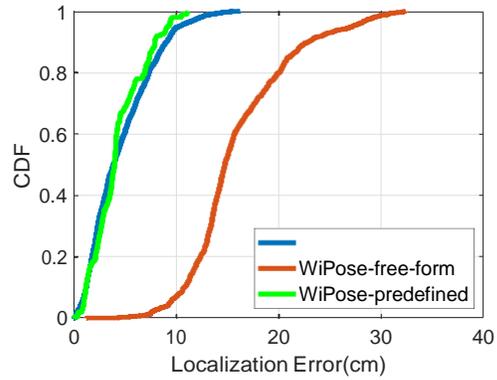
Fig. 18. Performance comparison for free-form activities only, and predefined activities only.

We first evaluate the overall performance of our system with the activities including both free-from and predefined activities that are described in the experimental setup. Figure 17 shows the cumulative distribution function (CDF) of tracking errors for both our system and WiPose. We can observe that the median tacking error of our system and WiPose is 3.9cm and 9.2cm, respectively. Further, we observe that 80 percentile tracking error is at around 7cm for our system, while it is more than 15cm for WiPose. The results demonstrate that our system has high tracking accuracy, whereas WiPose has worse performance primarily due to the activities under evaluation include free-form activities that WiPose cannot work well.

Moreover, Table 2 shows the average joint localization error for each joint as well as the overall result for all 8 joints. The range of joint localization error of our system is from 4.1cm to 5.1cm, whereas it is from 7.8cm to 15.4cm for WiPose. The overall localization error of our system is 4.6cm, while it is 10.1cm for WiPose. These results show that our system outperforms WiPose as our system is designed to work with free-form activities whereas WiPose is limited to a set of predefined activities.

We further compare the performance of our proposed system with WiPsoe under the free-form activity only (i.e., WiPose-free-form) and under the predefined activity only (i.e., WiPose-predefined), respectively. For WiPose, free-form activities are not seen during the training phase, whereas WiPose has seen and trained each predefined activity during the training phase. Figure 18 shows that our system has almost the same performance as that of WiPose-predefined. In particular, both have a median and 80 percentile localization error at around 3.9cm and 7.0cm, respectively. This shows that our system without specific training of predefined activities can work as well as the WiPose that with the training of the predefined activities. More, our system performs much better for the free-form activity as WiPose-free-form degrades significantly from 4.0cm to 14.7cm for median error and from 7.0cm to 20cm for 80 percentile error.

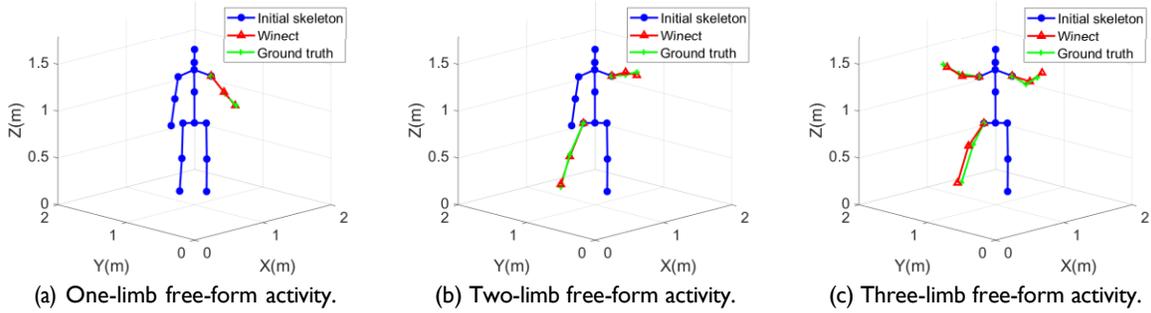

(a) One-limb free-form activity.   (b) Two-limb free-form activity.   (c) Three-limb free-form activity.

Fig. 19. Examples of constructed 3D human skeleton for multi-limb free-form activities.

Similar results can be observed from Table 2, which illustrates the tracking error for each joint. From Table 2, we can see the average joint localization error of WiPose-predefined is 4.3cm. The joint localization errors of WiPose-predefined range from 2.1cm to 6.1cm. Such a performance of WiPose-predefined is similar to that of our system. However, WiPose-free-form has large joint localization errors, which range from 12cm to 20.6cm. The reason is that when target activities are free-form and thus most of the target activities are not seen during the training for WiPose. Our system can accurately track free-form activities accurately since we leverage both movement signal separation and joint decomposition model. The above results demonstrate that our proposed system achieves comparable performance with WiPose-predefined and is able to simultaneously track free-form activities of multiple limbs with high accuracy.

To intuitively observe the performance of Winect, Figure 19 shows examples of constructed 3D skeletons of free-form activities. We color the initial skeleton with blue, the predicted skeleton with red and the ground truth with green. Figure 19(a), (b) and (c) illustrate examples for one-limb, two-limb and three-limb free-form activities. We can observe that 3D skeletons constructed by Winect are almost the same as the ground truths.

### 4.3 Impact of Different Number of Moving Limbs

Next, we study the impact of the different number of moving limbs by performing activities with one limb (i.e., one arm or one leg), two limbs (i.e., one arm and one leg, or two arms) and three limbs (i.e., two arms and one leg), respectively. Figure 20 illustrates the average localization errors for the different numbers of moving limbs which are 3.7cm, 4.6cm and 6.7cm for one limb, two limbs and three limbs, respectively. When the number of moving limbs decreases, the localization error will also decrease. This is because fewer moving limbs result in less complex reflection signals. Thus, it is easier to separate signals. Also, we can observe that our system can maintain relatively small errors even when three limbs are moving simultaneously. In practice, people only move one or two limbs at the same time in most cases. Hence, we can conclude that our system can track the activities of multiple limbs simultaneously with high accuracy.

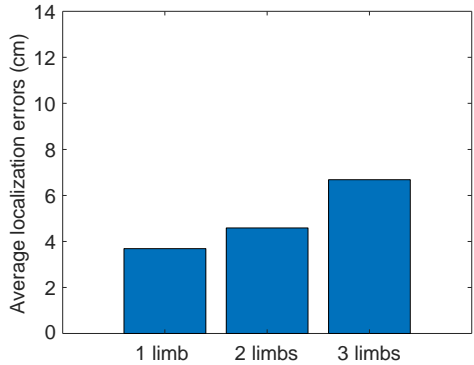
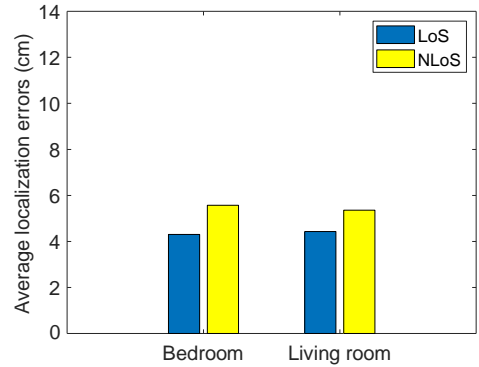

Fig. 20. System performance for different numbers of moving limbs.

Fig. 21. System performance under both NLoS and LoS scenarios.

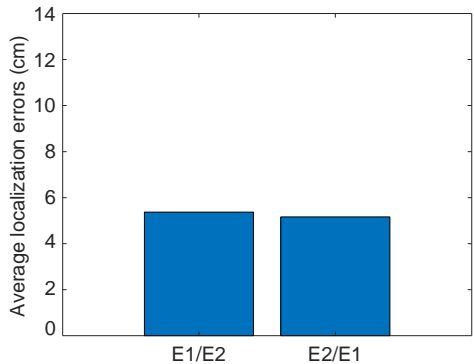
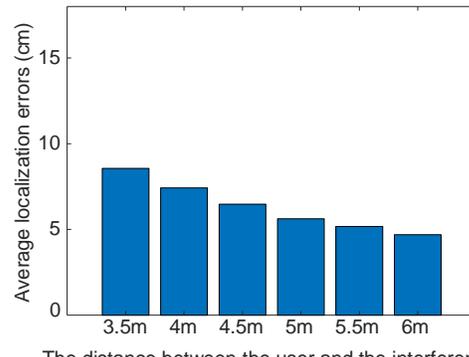

Fig. 22. System performance across different environments.

Fig. 23. System performance under ambient interferences.

### 4.4 Impact of Non-Line-of-Sight

We then discuss the impact of non-line-of-sight (NLoS) by placing a wooden screen between the person and each pair of transceivers in both the bedroom and the living room. Figure 21 presents the performance comparison under the NLoS and LoS scenarios in both two environments. Results show that the NLoS scenario slightly degrades the system performance. However, the average localization errors of NLoS are less than 6cm for both the bedroom and the living room. The results demonstrate that the proposed system could work in the NLoS scenario with relatively weak WiFi signals. This allows us to deploy the proposed system to a wider range of applications than computer vision-based systems.

### 4.5 Impact of Different Environments

Users could build the joint decomposition model of the system in one environment and apply it to a different environment. Hence, we study the impact of different environments on system performance. Specifically, we collect CSI data and build the joint decomposition model in the $1^{st}$ environment (i.e., bedroom) and test the model in the 2nd environment (i.e., living room). For simplicity, we denote this scenario as E1/E2. Similarly, we also evaluate the E2/E1 scenario. Figure 22 shows the average error across different environments. The overall cross-environment error is 5.3cm which is close to the overall same environment error which is 4.6cm. The reason is that we perform the static environment signal removal during the signal separation. In addition, the joint decomposition model does not rely on the background environment. The results clearly demonstrate that our system is environment-independent. This allows us to easily deploy the proposed system in various environments.

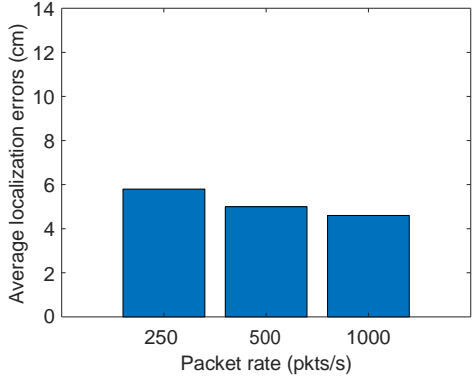
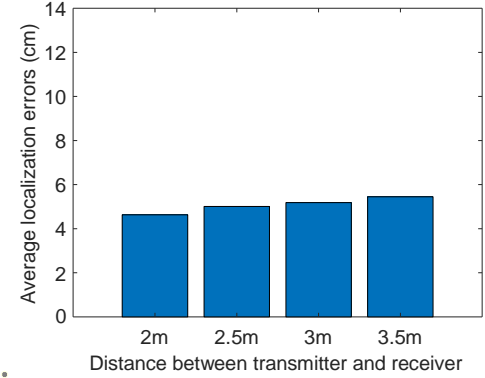

Fig. 24. System performance under different packet rates.   Fig. 25. System performance under different distances.

### 4.6 Robustness to Ambient Interference

In all of the evaluations described so far, we conduct experiments in static environments. A more realistic situation might be when a user is performing an activity, other people are walking or performing random movements nearby that can be considered as interferences. To study the robustness of our system to these ambient interferences, we vary the distance between the user and the interferences from 3.5m to 6m at a step size of 0.5m and evaluate the performance of Winect. As shown in Figure 23, when the distance between the user and the interferences is increased from 3.5m to 6m, the average localization error is reduced from 8.6cm to 4.7cm. This is because the farther the distance between the user and the interference is, the less disturbance the signal suffers. The results show that our system is robust to ambient interferences to a certain degree.

### 4.7 Impact of Packet Rate

We utilize a packet rate of 1000 packets per second in most of the experiments. However, the packet rate can be much lower in practice. To evaluate the impact of packet rate on the system performance, we down-sample the CSI data to 500 pkts/s and 250 pkts/s, respectively. Also, the corresponding segment length of CSI is decreased to 50 and 25, respectively. The results are shown in Figure 24. We can find that a higher packet rate leads to higher accuracy. Also, we can observe that the decrease of the packet rate has little impact on the performance of our system. For example, when the packet rate is only 250 pkts/s, the average error is still less than 5.8cm. Therefore, the minimum packet rate we recommend in our system is 250 pkts/s.

### 4.8 Impact of Different Distances

Next, we study the impact of the distance between the transmitter and receivers on the performance of the proposed system. As the height of the room (i.e., $z$-axis) remains constant, we change the distance between the transmitter and the receiver in $xy$-plane. The sensing area in the $xy$-plane is a square and the user always stands in the center of the area. We set the distance between the transmitter and the receiver as 2m, 2.5m, 3m and 3.5m on both the $x$-axis and $y$-axis. Figure 25 shows that when the distance is reduced from 3.5m to 2m, the error also reduces from 5.4cm to 4.6cm. This is because a shorter transmission distance leads to higher received signal strength. Thus, we could deploy receivers closer to the transmitter to have a better system performance. In a nutshell, the results indicate that our system works well in a typical room with a variety of distances.

### 4.9 Impact of Multiple Users

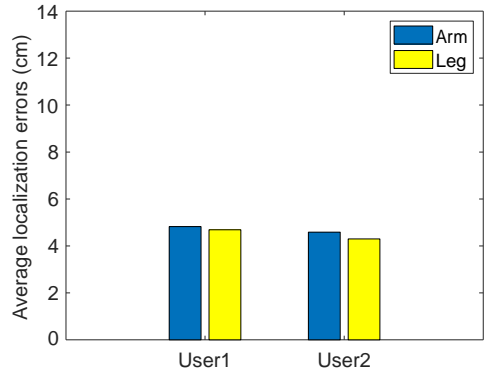

Fig. 26. System performance for multiple users.

In practice, there can be multiple users perform activities at the same time. Therefore, we evaluate the impact of multiple users by asking two users to move their limbs simultaneously. Figure 26 shows the average errors for both user1 and user2. In particular, for user1, the errors of the arm and leg are 4.8cm and 4.7cm, respectively. For user2, the errors of the arm and leg are 4.6cm and 4.3cm, respectively. We find that the errors slightly increase compared with the single-user scenario. This is because the increment of the number of users will increase the complexity of the reflected signal. Also, the tracking accuracy of legs is higher since the reflection area of the legs is larger and reflected signals are stronger. Because a typical room only allows two people to perform the activity without overlapping, we no longer increase the number of users. The results show that our system is able to accurately track the activities of multiple users simultaneously.

### 4.10 Performance of Limb Identification

One important step in our system is to identify the moving limbs. Therefore, we should evaluate the performance of identifications of which limb is moving and the number of moving limbs. We randomly select multiple (e.g., 100 in this work) CSI data to conduct the one-time identification. Figure 27(a) shows that our system achieves an overall accuracy of 98.5% for one-time identification of moving limbs. As shown in Figure 27(b), the overall accuracy of one-time identification of the number of moving limbs is over 98.6%. we can refine the results by applying multiple identifications with the majority rule. Figure 28(a) and (b) show that identifications of limbs and the numbers both achieve the accuracy close to 1 using three-time identification with the majority rule. Note a single identification only need 100 CSI packets which take 0.1s. We allow 50% overlap, thus a three-time identification with the majority rule only needs 0.2s. It usually takes a few seconds to perform one daily activity which is enough for limb identification. This evaluation demonstrates that our system can accurately identify multiple moving limbs and the total number of moving limbs.

|  | Left Arm | Right Arm | Left Leg | Right Leg |
|---|---|---|---|---|
| Left Arm | 0.98 | 0.01 | 0.01 | 0.00 |
| Right Arm | 0.02 | 0.97 | 0.00 | 0.01 |
| Left Leg | 0.00 | 0.00 | 1.00 | 0.00 |
| Right Leg | 0.00 | 0.00 | 0.01 | 0.99 |

(a) Performance of one-time identification for moving limbs.

|  | 1 Limb | 2 Limbs | 3 Limbs |
|---|---|---|---|
| 1 Limb | 1.00 | 0.00 | 0.00 |
| 2 Limbs | 0.01 | 0.99 | 0.00 |
| 3 Limbs | 0.01 | 0.02 | 0.97 |

(b) Performance of one-time identification for the number of moving limbs.

Fig. 27. Performance of one-time identification.

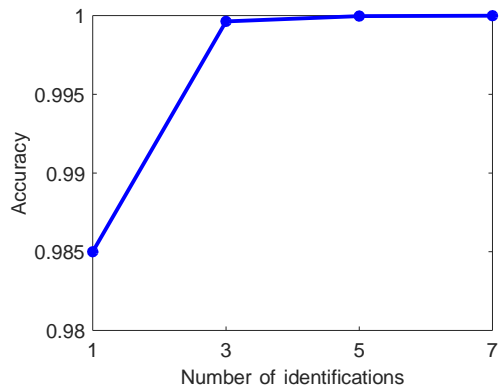

(a) Performance of multiple identifications with majority rule for moving limbs.

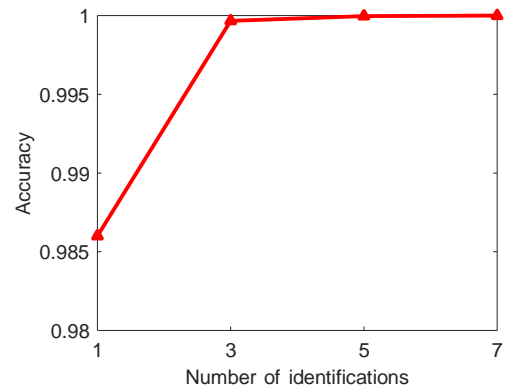

(b) Performance of multiple identifications with majority rule for the number of moving limbs.

Fig. 28. Performance of multiple identifications with majority rule.

## 5 DISCUSSION

Although Winect can track 3D free-form activities with high accuracy in different scenarios, it still has some limitations.

**Tracking in crowded environments.** In our experimental setup, each commodity WiFi device only has three antennas. Due to the limited number of antennas, our system can only track a small number of people simultaneously. However, there are many crowded environments with a large number of people, such as a theater and a bus station. Such crowded environments include complex scenes, such as occlusion, overlapping, and a large number of limbs. Our system thus is difficult to track each person in crowded environments with a limited number of antennas and WiFi devices. A potential solution is to improve the spatial resolution of the system by employing a general-purpose antenna extension [67] on the WiFi devices. Also, we may deploy more devices at different locations and directions to capture each person from different angles so as to track in crowded environments.

**Human positions and orientations.** As our system is environment-independent, the user can choose any position or orientation to perform free-form activities. This is because the signal separation and joint decomposition model of Winect are not affected by positions or orientations. In our evaluation, however, the user is asked to face the AoA sensor (i.e., the WiFi device used for deriving 2D AoA). This is because we only used one sensor to derive 2D AoA for the moving limb identification. If we deploy three AoA sensors in orthogonal directions, the user is free to change to any orientations. Still, our system might not work well when the user is walking while performing activities. The reason is that the signal reflection separation could be too complicated to resolve under the walking scenario. In order to achieve the tracking of poses for walking users, we could utilize the region proposal network [43] to generate possible person regions each time for pose estimation.

**Captioning human activity.** Our current system can only track free-form activities by estimating 3D skeletons. The proposed system cannot caption human activities (i.e., create a textual description of people's activities) as it is not designed to understand the contents of activities. However, captioning activity has found many applications such as human-computer interaction, smart home, and surveillance. In future work, we may combine the estimated human skeleton and the deep neural network to caption human activities and even interactions with objects or other people.

## 6 CONCLUSION

This paper presents Winect, which is capable of tracking 3D human poses for free-form activities. The proposed system does not rely on a set of predefined activities and can track free-form motions of multiple limbs simultaneously by reusing existing commodity WiFi devices. Our system enables 3D skeleton pose estimation by separating the entangled signals from different limbs and then modeling the relationships between the movements of the limb and the corresponding joints. Extensive experiments show that Winect can track 3D human pose with an average error of 4.6cm for various free-form activities and works well under various environments or scenarios including the none-line-of-sight (NLoS) scenarios.

## ACKNOWLEDGMENTS

We thank the anonymous reviewers for their insightful feedback. This work was partially supported by the NSF Grants CNS-2131143, CNS-1910519, and DGE-1565215.